\documentclass[sn-aps]{sn-jnl}

\usepackage{graphicx}%
\usepackage{multirow}%
\usepackage{amsmath,amssymb,amsfonts}%
\usepackage{amsthm}%
\usepackage{mathrsfs}%
\usepackage[title]{appendix}%
\usepackage{xcolor}%
\usepackage{textcomp}%
\usepackage{manyfoot}%
\usepackage{booktabs}%
\usepackage{algorithm}%
\usepackage{algorithmicx}%
\usepackage{algpseudocode}%
\usepackage{listings}%
\usepackage{algpseudocode,algorithm}
\usepackage{soul}
\usepackage{dsfont}
\usepackage{tabularx}
\usepackage{fancyhdr}
\fancypagestyle{firstpage}{
  \fancyhf{} 
  \rfoot{} 
  \cfoot{} 
  \lfoot{Published in \textit{Knowledge-Based Systems} (2022). \href{https://doi.org/10.1016/j.knosys.2022.109664}{https://doi.org/10.1016/j.knosys.2022.109664}} 
}

\begin{document}

\title[Article Title]{Stream-based active learning with linear models}


\author*[1,2]{\fnm{Davide} \sur{Cacciarelli}}\email{dcac@dtu.dk}

\author[1,3]{\fnm{Murat} \sur{Kulahci}}\email{muku@dtu.dk}

\author[2]{\fnm{John Sølve} \sur{Tyssedal}}\email{john.tyssedal@ntnu.no }

\affil[1]{\orgdiv{Department of Applied Mathematics and Computer Science}, \orgname{Technical University of Denmark}, \orgaddress{\city{Kgs. Lyngby}, \country{Denmark}}}

\affil[2]{\orgdiv{Department of Mathematical Sciences}, \orgname{Norwegian University of Science and Technology}, \orgaddress{\city{Trondheim}, \country{Norway}}}

\affil[3]{\orgdiv{Department of Business Administration, Technology and Social Sciences}, \orgname{Luleå University of Technology}, \orgaddress{\city{Luleå}, \country{Sweden}}}


\abstract{The proliferation of automated data collection schemes and the advances in sensorics are increasing the amount of data we are able to monitor in real-time. However, given the high annotation costs and the time required by quality inspections, data is often available in an unlabeled form. This is fostering the use of active learning for the development of soft sensors and predictive models. In production, instead of performing random inspections to obtain product information, labels are collected by evaluating the information content of the unlabeled data. Several query strategy frameworks for regression have been proposed in the literature but most of the focus has been dedicated to the static pool-based scenario. In this work, we propose a new strategy for the stream-based scenario, where instances are sequentially offered to the learner, which must instantaneously decide whether to perform the quality check to obtain the label or discard the instance. The approach is inspired by the optimal experimental design theory and the iterative aspect of the decision-making process is tackled by setting a threshold on the informativeness of the unlabeled data points. The proposed approach is evaluated using numerical simulations and the Tennessee Eastman Process simulator. The results confirm that selecting the examples suggested by the proposed algorithm allows for a faster reduction in the prediction error.}





\maketitle
\thispagestyle{firstpage} 


\section{Introduction} \label{sec:introduction}
The term big data seems to be ubiquitous in many fields of application, and industrial production is no different. However, in production, this can be somewhat misleading as it often refers to process data that is obtained through automated data collection schemes with minimal manual interference. Product-related data is usually scarcer particularly in high-volume manufacturing due to costs of inspection. This creates an imbalance in the amount of available data that can at times be quite substantial. Yet in many cases, predictive modeling relating process variables to product characteristics is sought after. Therefore, it will be beneficial to guide the data collection schemes for product characteristics through a real-time sampling methodology. In current production environments, sampling of the product characteristics is often performed at regular time intervals or at random. However, this approach can be ineffective as the informativeness of the observations at the time of sampling is not taken into account. This problem is reinforcing the interest of researchers and practitioners in active learning. Active learning-based sampling schemes use an instance selection criterion to strategically select data points that allow a faster reduction of the generalization error \cite{Kumar2020}. Over the last decades, many active learning approaches have been proposed, but most of the focus has been dedicated to the pool-based scenario \cite{Settles2009}. Pool-based active learning refers to a circumstance in which a large amount of unlabeled data is collected all at once and made available to the learner, which can then select offline the data points to be labeled with a greedy approach \cite{Lewis1994}.

In real-time applications for high-volume production, where samples are processed at a fast pace, evaluating all the available instances before making a choice might not be realistic. In these cases, the learner might only have a short time frame to make the sampling decision. Indeed, if a sample is not selected for the quality check, it might get lost in the downstream process and no longer be traceable. This is particularly relevant in high-volume production, where tracing individual parts is a challenge. Also, in a chemical process, we might not be able to measure the level of the variable of interest once a component undergoes a specific treatment. In these contexts, a much more sensible scenario is represented by stream-based active learning, which is sometimes referred to as selective sampling \cite{Cohn1994}. Stream-based active learning investigates a scenario where instances are processed one at a time and the learner has to determine immediately whether to keep the instance and query its label or discard it. The task is very similar to the one described by a notorious statistical riddle, the secretary problem \cite{Freeman1983}, where an observer sequentially interviews a certain number of applicants and, after each interview, a decision must be made on whether the applicant is hired or not. An exhaustive survey about stream-based active learning has been proposed by Lughofer \cite{Lughofer2017}, who classified existing online active learning methods by taking into account the data processing functionality, the model class (regression or classification), and many other relevant properties. The survey reveals how stream-based active learning methods have been mostly developed in the classification field. Regression models, on the other hand, are extremely useful in the development of soft sensors for hard-to-measure process variables or in quality control problems where a product's characteristic is measured on a continuous scale. That is why active learning in conjunction with regression models is capturing the interest of many researchers \cite{Chan2018,Shi2018,Ge2014,Tang2018}.

In this paper, we focus on the use of linear regression models. These models are well suited for stream-based active learning as they can easily be trained on a small number of observations, being composed of a small number of parameters. This property is also very useful if we want to efficiently retrain the model each time the design is augmented by including an additional observation \cite{Maccio2016}. Moreover, despite recent advances in terms of interpretability for deep learning models, linear regression models are still amongst the most easily interpretable models. Indeed, their parameters offer a straightforward quantitative contribution of each specific feature, and their input features are directly derived from the empirical observations \cite{AlvarezMelis}. Besides the direct interpretation that comes from the signs and magnitudes of the coefficients, linear models can also be used to construct confidence intervals on the parameter estimates and variable selection can also be easily incorporated into such models \cite{Efron2004}. Recently, additional variable selection methods for linear regression models have been suggested by Zhang et al. \cite{Zhang2018EarlyStop}. Being able to offer a robust feature importance analysis is particularly important in industrial problems, where practitioners and engineers might need to timely intervene in specific parts or components of the process to ensure safety and operational efficiency. The simplicity of these models and the low number of parameters that require tuning is also beneficial to foster their adoption and use in applications. Finally, linear regression models allow us to build on the optimal experimental design theory and leverage the criteria that are typically used to design highly efficient experiments. Despite the focus of this paper being dedicated to linear models, nonlinear models proved to be extremely useful in a wide variety of applications. In particular, deep learning models are very effective in dealing with complex high-dimensional data to perform tasks like image recognition, shape extraction, and pose recovery \cite{Hong2015,Hong2014,Yu2022,Yu2015,Hong2019}.

In this work, we propose a novel strategy to perform stream-based active learning with linear models. Given the impossibility to rank observations in real-time, we provide an algorithm that only uses unlabeled data to set a threshold on the informativeness of data points. Unlabeled data is also exploited in a semi-supervised manner to increase the predictive performance \cite{Frumosu2018}. We show how the proposed approach outperforms random sampling and state-of-the-art methods.

The remainder of this paper is organized as follows. In Section \ref{sec:preliminaries}, we define some basic concepts and discuss related works focusing on active learning for regression. Section \ref{sec:methods} introduces the proposed sampling strategy. In Section \ref{sec:experiments} we test our approach using numerical simulations; the Tennessee Eastman Process data is also used to evaluate its performance on a typical industrial process. Finally, Section \ref{sec:end} provides some conclusions.

\section{Preliminaries} \label{sec:preliminaries}
The active learning problem is defined by an imbalance between the availability of process variables $\mathbf{x} \in \mathbb{R}^p$ and the corresponding labels $y \in \mathbb{R}$. In many circumstances, industrial processes are characterized by the presence of easy-to-measure process variables, which are collected through automated collection schemes, and hard-to-measure variables, whose values are difficult to track during routine operations. Large plants, measurement delays, and environments hostile to the survival of measuring devices are all situations where hard-to-measure process variables are commonly encountered \cite{Fortuna2007}. Similar situations can be addressed by utilizing soft sensors based on predictive models to forecast the true values of hard-to-measure variables. For modeling purposes, we assume that the true underlying relationship between the process variables and the product information or hard-to-measure variable can be expressed with a linear model of the form

\begin{equation}
    \mathbf{y}=\mathbf{X} \boldsymbol{\beta}+\boldsymbol{\epsilon}
\end{equation}

\noindent
where
\begin{equation*}
    \mathbf{y}=\left[\begin{array}{c}
y_1 \\
y_2 \\
\vdots \\
y_n
\end{array}\right], \quad \mathbf{X}=\left[\begin{array}{cccc}
x_{11} & x_{12} & \cdots & x_{1 p} \\
x_{21} & x_{22} & \cdots & x_{2 p} \\
\vdots & \vdots & & \vdots \\
x_{n 1} & x_{n 2} & \cdots & x_{n p}
\end{array}\right], \quad \boldsymbol{\beta}=\left[\begin{array}{c}
\beta_1 \\
\beta_2 \\
\vdots \\
\beta_p
\end{array}\right], \quad \text { and } \quad \boldsymbol{\epsilon}=\left[\begin{array}{c}
\epsilon_1 \\
\epsilon_2 \\
\vdots \\
\epsilon_n
\end{array}\right]
\end{equation*}

\noindent
$\mathbf{y}$ is a $n \times 1$ vector of response variables, $\mathbf{X}$ is a $n \times p$ model matrix, $\boldsymbol{\beta}$ is a $p \times 1$ vector of regression coefficients, and $\boldsymbol{\epsilon}$ is a $n \times 1$ vector representing the noise, with covariance matrix $\sigma^2 \mathbf{I}$. Here $n$ represents the total number of observations and $p$ the number of process variables (as well as the number of parameters in a model with main effects only and no intercept). If the predictors and the response are not centered, an intercept term may be added to the model. In that case, the size of the model matrix becomes $n \times (p+1)$, and $\boldsymbol{\beta}$ a $(p+1) \times 1$ vector. When $k \geq p$ observations are available to the learner, we can obtain a least squares estimator for $\boldsymbol{\beta}$ using

\begin{equation} \label{eq:estimation}
    \widehat{\boldsymbol{\beta}}=\left(\mathbf{X}^\top \mathbf{X}\right)^{-1} \mathbf{X}^\top \mathbf{y}
\end{equation}

\noindent
such that the fitted linear regression model will be given by $\widehat{\mathbf{y}}=\mathbf{X} \widehat{\boldsymbol{\beta}}$ and its residuals by $\mathbf{e}=\mathbf{y}-\widehat{\mathbf{y}}$. A key distinction between the experimental design approach and stream-based active learning concerns the assumption we make about the randomness of the process variables. In design of experiments, the $\mathbf{x}$ vectors are assumed to be fixed while in this case we assume that $\mathbf{X}$ is composed by random vectors, as the individual observations are sampled from a process subject to random variation and we are not able to set the precise location of the incoming data points. However, conditional on the observed $\mathbf{X}$ variables, $\left(\mathbf{x}_1, \ldots, \mathbf{x}_p\right)$, Equation \ref{eq:estimation} still applies. It should be noted that the coefficients $\widehat{\boldsymbol{\beta}}$ determined using Equation \ref{eq:estimation} may not be stable if the data matrix $\mathbf{X}$ is affected by multicollinearity. To deal with this issue and achieve robust results, a solution might be to use a ridge estimate for the coefficients, $\widehat{\boldsymbol{\beta}}_{\text {ridge }}=\left(\mathbf{X}^\top \mathbf{X}+\lambda \mathbf{I}\right)^{-1} \mathbf{X}^\top \mathbf{y}$. An alternative approach to tackle multicollinearity is to perform a pre-whitening of $\mathbf{X}$ to remove the dependencies between the components.

We assume a small, labeled training set is initially available and can be used to fit the first regression model, as is common practice in active learning applications \cite{Burbidge2007,Ge2014,Ge2016}. The number of observations provided to the learner usually corresponds to a modest fraction (e.g., 5\%) of the total number of instances available \cite{Reyes2018,Cai2013}. After the first model has been built, the learner is granted a certain operational budget $b$ to augment the design matrix by including additional observations. Some approaches focus on this problem in a pool-based context, in which the total number of observations $n$ is represented by a closed and static set $\mathcal{U}$ and the label of a specific data point can always be queried. Among these approaches, query-by-committee (QBC) \cite{Burbidge2007} suggests building an ensemble of regression models trained on bootstrap replica of the original training set. Once the ensemble, or committee, has been built, the variance of the predictions made by the committee members is computed for each unlabeled observation $\mathbf{x} \in \mathcal{U}$. This metric, also referred to as ambiguity, is used to rank the instances belonging to the unlabeled set $\mathcal{U}$ by prioritizing the data points with the highest variance. Expected model change maximization (EMCM) \cite{Cai2013} is another noteworthy study that focuses on the observations that impact the most the current model’s parameters. The model change is defined as the difference between the current model parameters and the parameters obtained after fitting the model on the augmented design, including the unlabeled observation $\mathbf{x} \in \mathcal{U}$ that is currently under evaluation. Because the learner does not have access to the true label for that data point, it estimates it using the mean prediction of a bootstrap ensemble, as the one employed by QBC. Another offline approach, inspired by statistical process control, combines the Hotelling $T^2$ statistic and the squared prediction error of a principal component regression (PCR) model to obtain a sampling evaluation index \cite{Ge2014}.
Besides the fact that all these methods focus on the pool-based scenario, it should be noted that the approaches that use ensembles may not be well suited for the online scenario, given the higher computational cost associated with training and updating the models.

Optimal experimental theory is another field of research that is intrinsically related to active learning \cite{Karlin1966,Myers2016}. Optimal designs aim to reduce the cost of experimentation by proposing design matrices that allow a robust parameter estimation with the minimum number of runs. The most commonly employed optimality criteria are D-optimality \cite{John1975} and A-optimality. Important properties of a design can be derived from the moment matrix, or information matrix, which is defined as

\begin{equation} \label{eq:moment}
    \mathbf{M}=\frac{\mathbf{X}^\top \mathbf{X}}{N}
\end{equation}

\noindent
where $N$ represents the total number of runs. The moment matrix specifies the distribution of points in space and can be used to describe the design geometry. In a $2^k$ factorial design, where variables are expressed in coded units $(-1, +1)$, the moment matrix is equal to the identity matrix $\mathbf{I}_{k}$, as the columns of the design are orthogonal. In an orthogonal design, all the parameters can be estimated independently of one another \cite{Montgomery2012}. D-optimal designs try to pursue such property by focusing on good model parameter estimation. Inverting the moment matrix we obtain the scaled dispersion matrix given by

\begin{equation}
    \mathbf{M}^{-1}=N\left(\mathbf{X}^\top \mathbf{X}\right)^{-1}
\end{equation}

\noindent
This matrix contains the variances and covariances of the estimated coefficients of the regression model, scaled by $N / \sigma^2$  \cite{Myers2016}. Indeed, if the $k$ observations used to estimate $\widehat{\boldsymbol{\beta}}$ are i.i.d. and $\boldsymbol{\epsilon} \sim \mathcal{N}\left(\mathbf{0}, \sigma^2 \mathbf{I}\right)$, we have

\begin{equation}
    \widehat{\boldsymbol{\beta}}_{k} \mid \mathbf{X} \sim \mathcal{N}\left(\boldsymbol{\beta},\left(\mathbf{X}^\top \mathbf{X}\right)^{-1} \sigma^2\right)
\end{equation}

\noindent
It can be demonstrated how by increasing the determinant of $\mathbf{M}$, the variances and covariances of the model coefficients are reduced, leading to a better estimation of $\boldsymbol{\beta}$. A D-optimal design is attained by maximizing the determinant of the moment matrix. Formally, we are seeking the design $\mathcal{D}^*$ that satisfies

\begin{equation}
    \max _{\mathcal{D}}|\mathbf{M}(\mathcal{D})|=\left|\mathbf{M}\left(\mathcal{D}^*\right)\right|
\end{equation}

\noindent
A-optimality is another important optimality criterion that tries to achieve good parameter estimation by minimizing the sum of the individual variances of the coefficients. This is achieved by the design  $\mathcal{D}^*$ that satisfies

\begin{equation}
    \min _{\mathcal{D}} \operatorname{tr}[\mathbf{M}(\mathcal{D})]^{-1}=\operatorname{tr}\left[\mathbf{M}\left(\mathcal{D}^*\right)\right]^{-1}
\end{equation}

\noindent
as the variances of the coefficients can be found on the diagonal of the scaled dispersion matrix multiplied by $N / \sigma^2$. It should be noted that A-optimality does not consider the covariances between coefficients. 

Recently, the concept of A-optimality has been extended to stream-based active learning \cite{Fontaine2021,Riquelme2017AAAI}. That is, the approach has been extended outside the design of experiments framework, assuming $\mathbf{X}$ is composed of random vectors and the observations are sequentially drawn. Riquelme et al. \cite{Riquelme2017AAAI} show how to set a threshold to perform online active learning for linear regression models by minimizing the sum of the individual variances of $\widehat{\boldsymbol{\beta}}$. They state that, in order to achieve A-optimality and minimize the trace of the inversed scaled dispersion matrix, the eigenvalues of the moment matrix should be as balanced as possible. This is because the eigenvalues of $\mathbf{X}^\top \mathbf{X}$ represent the trace of $\mathbf{X}^\top \mathbf{X}$, which is also given by the sum of the norm of the observations. For this reason, they propose a norm-thresholding algorithm that pursues A-optimality by selecting observations with large, scaled norm. The scaling step can be ignored when whitening is used to remove dependencies. Finally, the design is augmented with the observations $\mathbf{x}$ whose norm exceeds a threshold $\Gamma$ given by

\begin{equation} \label{eq:nt}
    \mathds{P}(\|\mathbf{x}\| \geq \Gamma)=\alpha
\end{equation}

\noindent
where $\alpha$ is the ratio of observations we are willing to label out of the incoming data stream. This value is strongly dependent on the budget $b$ and the sampling rate used to collect the data.

Another noteworthy approach focusing on stream-based active learning for regression tasks has been suggested by Lughofer and Pratama \cite{Lughofer2018}. In this paper, the authors propose a single-pass selection criterion that takes into account ignorance about the input space, uncertainty in predictive model outputs, and uncertainty in model parameters. The main difference with our approach is that Lughofer and Pratama focus on the use of Takagi-Sugeno (TS) fuzzy models \cite{Lughofer2011}, combining adaptive error bars for the model output and A-optimality for the variances of the estimated parameters. Conversely, our method relies on statistical linear regression and tries to combine the exploration of lesser-known input space regions with accurate parameter estimates by employing the idea of D-optimality.

\section{Proposed approach} \label{sec:methods}
In this work, we try to improve the approach proposed by Riquelme et al. \cite{Riquelme2017AAAI} by moving from A-optimality to D-optimality. We believe that taking into account the covariance between the estimates of the model coefficients might be particularly advantageous with large datasets and models, where many factors might be active and influence the response. To adapt the D-optimality criterion to stream-based active learning, we start from the connection between D-optimality and prediction variance (PV) highlighted by Myers et al. \cite{Myers2016}. The PV at a point $\mathbf{x}^{(m)}$ is the variance of the predictor $\widehat{\mathbf{y}}\left(\mathbf{x}^{(m)}\right)$, which corresponds to $\operatorname{Var}\left(\mathbf{x}^{(m)\top} \widehat{\boldsymbol{\beta}}\right)$, and is given by

\begin{equation}
    \operatorname{PV}(\mathbf{x})=\sigma^2 \mathbf{x}^{(m) \top}\left(\mathbf{X}^\top \mathbf{X}\right)^{-1} \mathbf{x}^{(m)}
\end{equation}

\noindent
where $$\mathbf{x}^{(m)}$$ represents the data point where the variance is being estimated, expanded to the model form. We can also express the variance in a scale-free form using the scaled prediction variance (SPV), which is computed as

\begin{equation}
    \operatorname{SPV}(\mathbf{x})=N \mathbf{x}^{(m) \top}\left(\mathbf{X}^\top \mathbf{X}\right)^{-1} \mathbf{x}^{(m)}
\end{equation}

\noindent
It should be noted that the SPV is a quadratic form of the inverse moment matrix $\mathbf{M}^{-1}$, as it can also be written as $\mathbf{x}^{(m) \top} \mathbf{M}^{-1} \mathbf{x}^{(m)}$. Since SPV considers the total number of runs $N$, it can be used to assess the quality of a design on a per observation basis. In the online scenario, we are not interested in comparing designs of different sizes but rather we investigate the individual contributions of incoming data points to the current design. In this circumstance, we can discard $N$ and use the unscaled prediction variance (UPV), which is calculated as

\begin{equation}
    \operatorname{UPV}(\mathbf{x})=\mathbf{x}^{(m) \top}\left(\mathbf{X}^\top \mathbf{X}\right)^{-1} \mathbf{x}^{(m)}
\end{equation}

\noindent
As anticipated in Section \ref{sec:preliminaries}, we are already given an initial random design that contains some labeled examples, which is being used to fit an initial model. Then, we are interested in augmenting our design by iteratively selecting observations from a continuous stream. Pursuing D-optimality, we aim at collecting observations that allow us to maximize the determinant of the moment matrix $\mathbf{M}$. If we consider that the current design is composed by $k$ observations, we can decompose the numerator of the moment matrix (Equation  \ref{eq:moment}) before the design is augmented by including the $(k+1)$th observation as

\begin{equation}
    \mathbf{X}_k^\top \mathbf{X}_k=\mathbf{X}_{k+1}^\top \mathbf{X}_{k+1}-\mathbf{x}_{k+1} \mathbf{x}_{k+1}^\top
\end{equation}

\noindent
we can then express the determinant of $\mathbf{X}_k^\top \mathbf{X}_k$ as the product of the determinant of the numerator of the augmented moment matrix and a second term as in

\begin{equation}
    |\mathbf{X}_k^\top \mathbf{X}_k|=
    |\mathbf{X}_{k+1}^\top \mathbf{X}_{k+1}-\mathbf{x}_{k+1} \mathbf{x}_{k+1}^\top|=
    |\mathbf{X}_{k+1}^\top \mathbf{X}_{k+1}||1-\mathbf{x}_{k+1}^\top(\mathbf{X}_{k+1}^\top \mathbf{X}_{k+1})^{-1} \mathbf{x}_{k+1}|
\end{equation}

\noindent
It should be noted that the second term of the above equation is a scalar, irrespective of the number of variables $p$ and the number of observations $k$. From there, we can observe that 

\begin{equation}
    \frac{|\mathbf{X}_{k+1}^\top \mathbf{X}_{k+1}|}{|\mathbf{X}_k^\top \mathbf{X}_k|}=\frac{1}{1-\mathbf{x}_{k+1}^\top(\mathbf{X}_{k+1}^\top \mathbf{X}_{k+1})^{-1} \mathbf{x}_{k+1}}
\end{equation}

\noindent
From the properties of the hat matrix, which is generally defined as $\mathbf{H}=\mathbf{X}\left(\mathbf{X}^\top \mathbf{X}\right)^{-1} \mathbf{X}^\top$, we know that $0 \leq h_{j j} \leq 1$ is true for each element $h_{j j}$ of $\mathbf{H}$ \cite{Hoaglin1978Hat}. It follows that $\mathbf{x}_{k+1}^\top\left(\mathbf{X}_{k+1}^\top \mathbf{X}_{k+1}\right)^{-1} \mathbf{x}_{k+1} \leq 1$. Hence, we can conclude that the determinant of the new, enlarged, training set is maximized by seeking observations $\mathbf{x}$ that maximize $\mathbf{X}_{k+1}^\top\left(\mathbf{X}_{k+1}^\top \mathbf{X}_{k+1}\right)^{-1} \mathbf{X}_{k+1}$. That is, we will only select points that maximize the UPV. This may be explained by the fact that a data point for which we have a large prediction variance represents a less known region of the input space, and the regression model will highly benefit from its inclusion in the design. From Myers et al. \cite{Myers2016} we have that maximizing $\mathbf{x}_{k+1}^\top\left(\mathbf{X}_{k+1}^\top \mathbf{X}_{k+1}\right)^{-1} \mathbf{x}_{k+1}$ is equivalent to maximizing $\mathbf{x}_{k+1}^\top\left(\mathbf{X}_k^\top \mathbf{X}_k\right)^{-1} \mathbf{x}_{k+1}$, which is the UPV using the fitted model before the new point has been added to the training set.

Finally, following the norm-thresholding approach, we can set an upper control limit on new observations as

\begin{equation} \label{eq:threshold}
    \mathds{P}\left(\mathbf{x}_{k+1}^\top\left(\mathbf{X}_k^\top \mathbf{X}_k\right)^{-1} \mathbf{x}_{k+1} \geq \Gamma\right)=\alpha
\end{equation}

\noindent
In practice, as suggested by Riquelme et al. \cite{Riquelme2017AAAI}, when we start to observe the data points coming from the process, we allocate a first initial set of points to estimate the distribution of $\mathbf{x}_{k+1}^\top\left(\mathbf{X}_k^\top \mathbf{X}_k\right)^{-1} \mathbf{x}_{k+1}$. In this work, we used kernel density estimation (KDE) with a Gaussian kernel. The initial set is also used to estimate the sample covariance matrix $\boldsymbol{\Sigma}$. By performing an eigenvalue decomposition we can then express $\boldsymbol{\Sigma}$ as $\mathbf{U} \boldsymbol{\Lambda} \mathbf{U}^\top$, where $\mathbf{U}$ is an orthogonal matrix, whose ith column corresponds to the $i$th eigenvector of $\boldsymbol{\Sigma}$, and  $\boldsymbol{\Lambda}$ is a diagonal matrix with the eigenvalues of $\boldsymbol{\Sigma}$ on the diagonal. The incoming observations $\mathbf{x}$ can then be whitened using

\begin{equation}
    \mathbf{z}=\boldsymbol{\Lambda}^{-1 / 2} \mathbf{U}^\top \mathbf{x}
\end{equation}

\noindent
Before the whitening step, data can be centered and scaled using the sample mean and variances obtained from the initial set. In industrial contexts, when a lot of unlabeled process data is available in the form of a historical database, this step can also be performed offline. In this case, by fitting a principal component analysis (PCA) model to the large unlabeled dataset and using it to transform the incoming observations, we could improve the predictive performance using a semi-supervised PCR as suggested by Frumosu and Kulahci \cite{Frumosu2018}. The use of semi-supervised classification models has also received some attention in active-learning problems \cite{He2017,Fernandes2020,Leng2013}. Indeed, semi-supervised learning and active learning are both techniques that deal with scarcity of labels. However, they do so in two different ways. With semi-supervised learning, we try to get the most out of the currently available unlabeled data, whereas with active learning we try to acquire new data in the most effective way. 

Algorithm \ref{alg:1} describes the complete stream-based active learning procedure with the proposed approach, which might also be referred to as conditional D-optimality (CDO).

\begin{algorithm}[h]
\caption{Stream-based active learning using CDO}
\label{alg:1}
\begin{algorithmic}[h]
\Require an initial random design $\mathbf{X}$; a data stream $\mathbf{S}$; a warm-up length $w$; a sampling rate $\alpha$; a budget $b$
\State Set $\mathbf{W} = \varnothing$ \Comment{warm-up set to estimate $\boldsymbol{\Sigma}$ and $\Gamma$}
\State $i \gets 1$, $c \gets 0$ \Comment{$c$ represents the currently used budget}
\While{$i \leq w$}
\State Observe the $i$th data point $\mathbf{x}_i \in \mathbf{S}$
\State Select $x_i$: $\mathbf{W} = \mathbf{W} \cup \mathbf{x}_i$
\State $i \gets i + 1$
\EndWhile
\State Estimate the covariance matrix $\boldsymbol{\Sigma}$ of $\mathbf{W}$ and perform eigendecomposition $\boldsymbol{\Sigma}$ as $\mathbf{U} \boldsymbol{\Lambda} \mathbf{U}^\top$
\State Whiten the initial design by computing $\mathbf{Z}=\boldsymbol{\Lambda}^{-1 / 2} \mathbf{U}^\top \mathbf{X}$
\State Whiten the warm-up observations by computing $\mathbf{V}=\boldsymbol{\Lambda}^{-1 / 2} \mathbf{U}^\top \mathbf{W}$
\State Estimate $\Gamma$ using KDE on $\mathbf{V}$ with the desired sampling rate $\alpha$ using Equation \ref{eq:threshold} with $\mathbf{Z}$ and $\mathbf{V}$
\While{$c \leq b$ and $i \leq |\mathbf{S}|$}
\State Observe the $i$th data point $\mathbf{x}_i \in \mathbf{S}$
\State Whiten $\mathbf{x}_i$ by computing $\mathbf{z}_i=\boldsymbol{\Lambda}^{-1 / 2} \mathbf{U}^\top \mathbf{x}_i$
\If{$\mathbf{z}_i^\top\left(\mathbf{Z}^\top \mathbf{Z}\right)^{-1} \mathbf{z}_i \geq \Gamma$}
\State Ask for the label $y_i$ and augment the labeled dataset: $\mathbf{Z}=\mathbf{Z} \cup \mathbf{z}_i$
\State $c \gets c + 1$ \Comment{pay for the label}
\State Update threshold $\Gamma$ to measure the UPV of the enlarged design
\Else
\State Discard $\mathbf{x}_i$
\EndIf
\State $i \gets i + 1$
\EndWhile
\end{algorithmic}
\end{algorithm}

\begin{figure}[h]
  \centering
  \includegraphics[width=.7\linewidth]{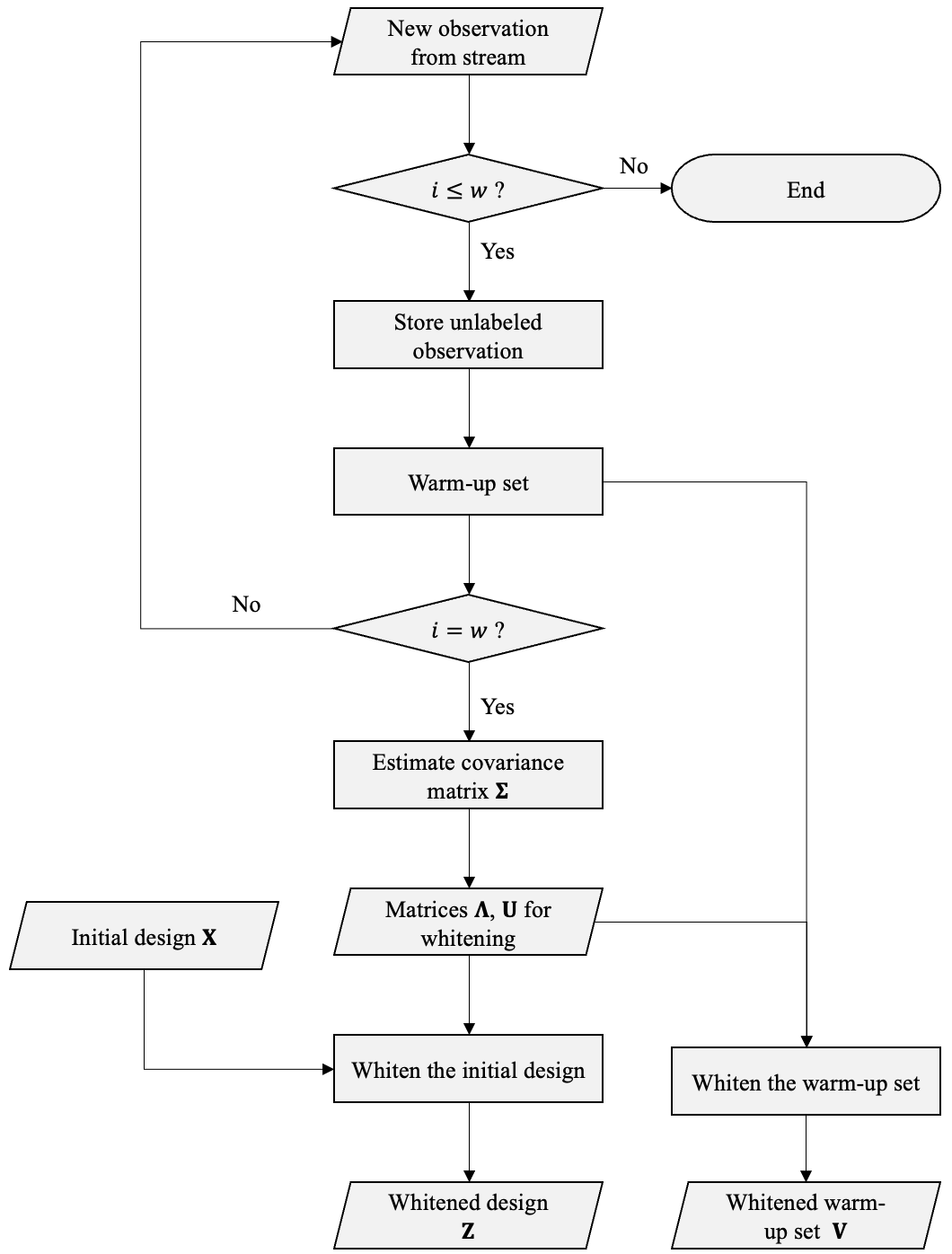}
  \caption{Flowchart of the warm-up phase of the stream-based active learning procedure.}
  \label{fig:fc1}
\end{figure}

\begin{figure}[h]
  \centering
  \includegraphics[width=.9\linewidth]{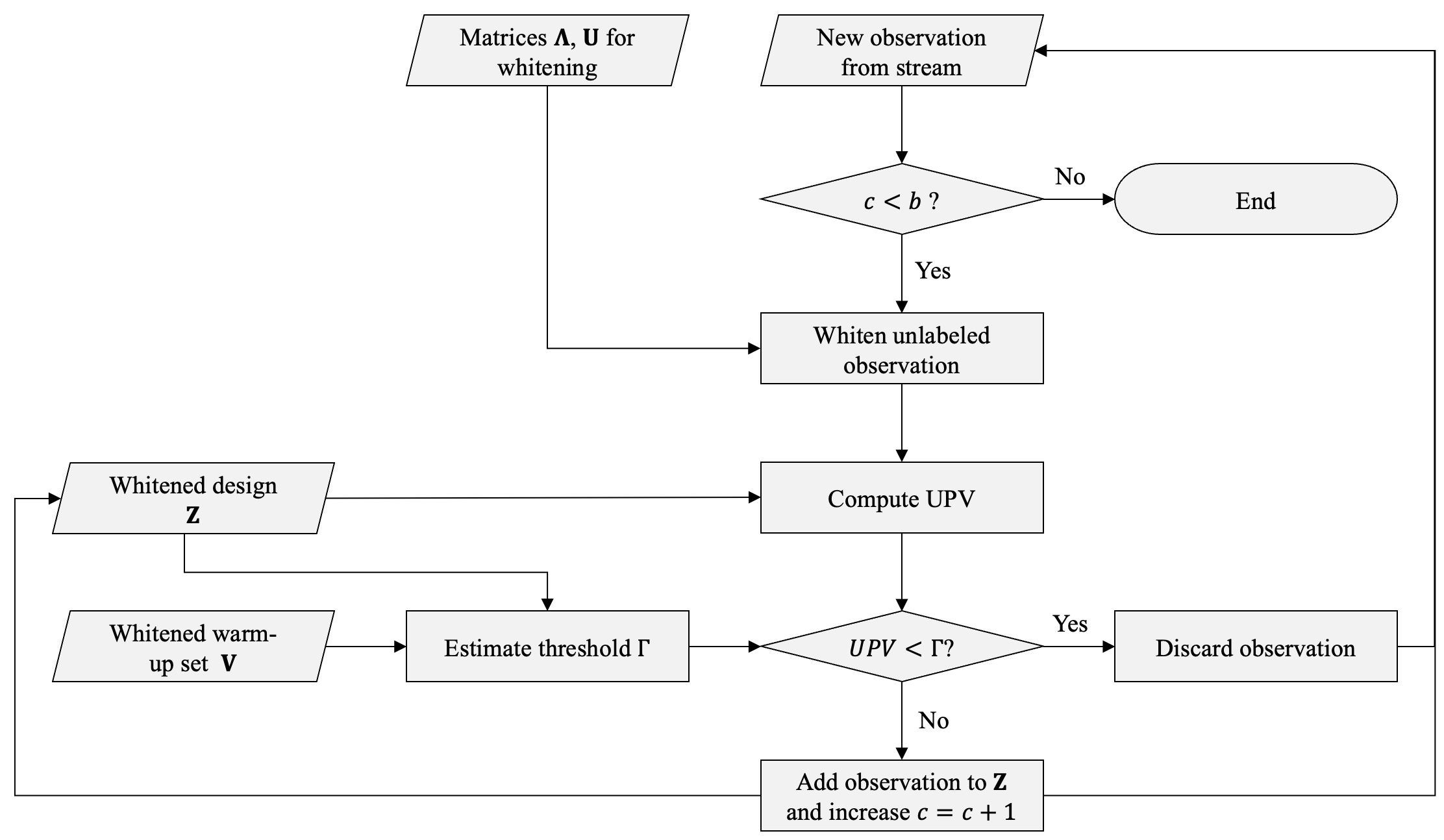}
  \caption{Flowchart of the instance selection phase of the stream-based active learning procedure.}
  \label{fig:fc2}
\end{figure}

An alternative representation of the CDO active learning routine is reported in the flowcharts in Figures \ref{fig:fc1} and \ref{fig:fc2}.  The first flowchart depicts the warm-up phase, which is represented by the first 10 steps of Algorithm \ref{alg:1}. The warm-up set is very important for the algorithm and serves two main purposes. First, it allows to estimate the covariance matrix of the data, which is later used for whitening the incoming observations. Secondly, it provides a set of unlabeled observations that can be leveraged to estimate the distribution of the UPV. The primary purpose of the whitening step is to address the multicollinearity issue in linear regression modeling, which can be aggravated when dealing with real-world data. The whitening step also ensures comparability with the norm-thresholding approach. Indeed, the norm-thresholding method without whitening would require computing a weighted norm to deal with dependencies between the components. The second flowchart represents the instance selection phase, the core of the active learning strategy. At this stage, we compute the UPV for the new observation sampled from the stream and we compare it to a pre-defined threshold. If the UPV computed at this point exceeds $\Gamma$, we query its label and include the labeled example in the training set. After the inclusion of the new point, a new threshold is estimated. The threshold is found by applying Equation \ref{eq:threshold} to the whitened warm-up set $\mathbf{V}$. That is, $\mathbf{X}_k$ is substituted by $\mathbf{Z}$, the currently labeled training set after whitening, and $\mathbf{x}_{k+1}$ is given by each unlabeled data point belonging to $\mathbf{V}$. By doing so, we obtain a one-dimensional array that has the same cardinality as the number of observations in $\mathbf{V}$. These statistics are then used to approximate the distribution of the UPV using KDE and determine the $\alpha$-upper percentile.

\section{Experiments} \label{sec:experiments}
In the experiments, we compare the proposed method to the norm-thresholding approach and random sampling. The methods are tested using numerical simulations and data from a chemical process simulator. All the approaches start from the same labeled training set and then they iteratively augment the design until the budget constraint b is met. The performance of the models is expressed, in predictive terms, by the root mean squared error (RMSE) of the predictions on a separate test set of $n$ observations

\begin{equation}
    \mathrm{RMSE}=\sqrt{\sum_{i=1}^n \frac{\left(\hat{y}_i-y_i\right)^2}{n}}
\end{equation}

\noindent

\subsection{Numerical simulations} \label{subsec:simulations}
To analyze the validity of the proposed method in the stream-based scenario, multiple datasets were created, each with a different dimensionality in terms of the number of process variables $p$. Within each dataset, incoming observations $\mathbf{x}$ are distributed according to a joint multivariate normal distribution $\mathcal{N}_p\left(\mathbf{0}, \boldsymbol{\Sigma}_0\right)$, where $\boldsymbol{\Sigma}_0$ is given by $\sigma^2 \mathbf{I}$, with $\sigma^2=1$. We ran 50 simulations for each number of  $p$ and, for each simulation run, the true coefficients are generated as $\beta \sim U(-5,5)$. It should be noted that $\beta$ has the same dimensionality as $\mathbf{x}$. This means that, using a first order model, a coefficient for each process variable needs to be estimated. The noise is given by $\epsilon \sim \mathcal{N}(0,1)$. For each scenario, an initial random design $\mathbf{X}$ is assumed available to the learner. We selected $p+2$ number of observations for the initial design, as  $k\geq p$  observations are needed to uniquely estimate $\widehat{\boldsymbol{\beta}}$.

\begin{figure}[h]
  \centering
  \includegraphics[width=\linewidth]{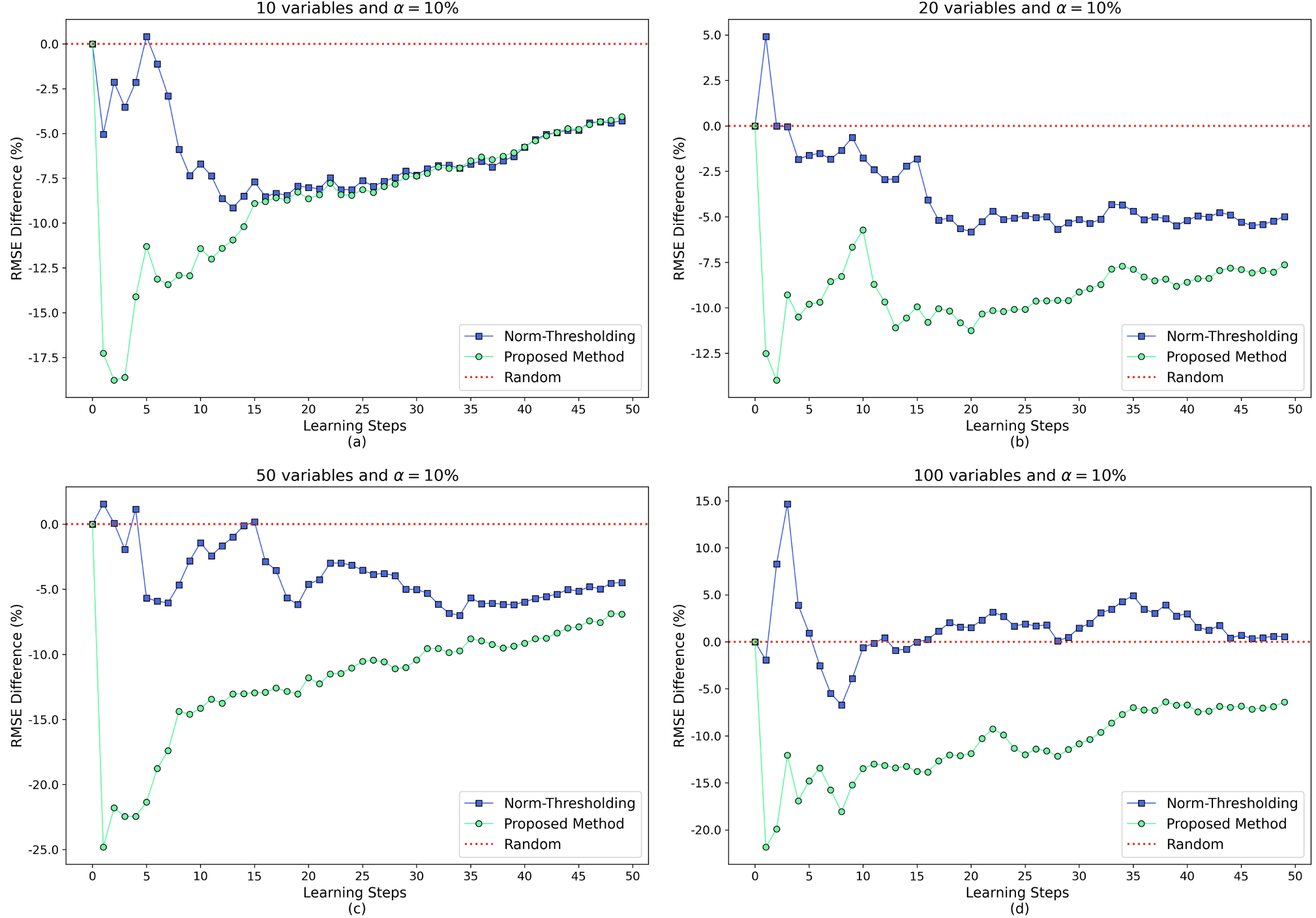}
  \caption{Percentage difference in RMSE between random sampling and the active learning methods, using $\alpha$=10\% (50 simulations).}
  \label{fig:numsim1}
\end{figure}

The learning curves reported in Figures \ref{fig:numsim1} and \ref{fig:numsim1} show the difference between the RMSE obtained with the two active learning strategies, using random sampling as the baseline. For each learning step, the percentage RMSE difference reported in the plots is obtained by computing $(\text {RMSE}_{\text {Active Learning}}-\text{RMSE}_{\text {Random}}) / \text{RMSE}_{\text {Random}} * 100$. This allows us to display a scale-free performance metric while comparing the different scenarios. 
The methods are tested using $b=50$ and with different levels for the $\alpha$ shown in Equations \ref{eq:nt}, and \ref{eq:threshold}. In the case of random sampling, $\alpha$ represents the probability of selecting an incoming observation. That is, each time a new sample arrives, a number $s \sim U(0,1)$ is generated and the data point is only selected if $s \geq 1-\alpha$. The warm-up length $w$ was set to 500 observations and it is being used by all the methods to estimate the covariance matrix, which is used for whitening the observations in a semi-supervised fashion. Moreover, it ensures comparability between the three strategies by setting the same starting points for the data streams. The models have been fitted without the intercept term as both process variables and outcome are centered.

\begin{figure}[h]
  \centering
  \includegraphics[width=\linewidth]{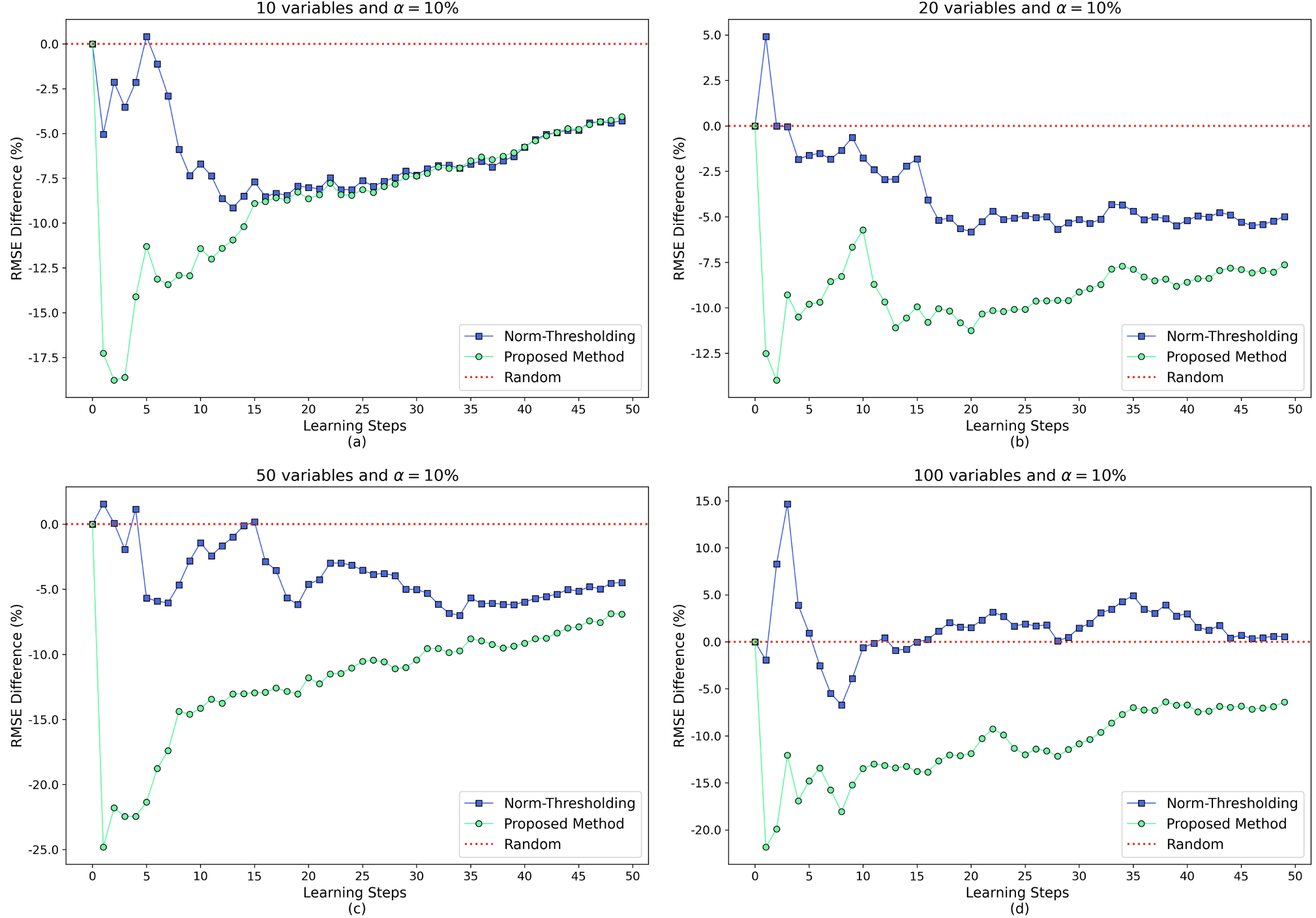}
  \caption{Percentage difference in RMSE between random sampling and the active learning methods, using $\alpha$=1\% (50 simulations).}
  \label{fig:numsim2}
\end{figure}

Figure \ref{fig:numsim1} shows the performance when using an $\alpha$ equal to 10\%. The x-axis reports the learning steps, which correspond to the inclusion of an additional observation to the training set. Indeed, when the design is augmented, the model is updated and new predictions are obtained for the same separate test set. It should be noted that the RMSE obtained in the first learning step is the same for the three methods, as all the models start from the same random design. It can be seen how the performance of the two active learning methods converges to the one obtained through random sampling as the number of labeled examples in the training set increases. Instead, when the number of labeled examples is lower, active learning proves to be particularly convenient. However, the proposed approach dominates the other strategies in all the scenarios.  Furthermore, it should be noted how the norm-thresholding algorithm seems to worsen when more and more parameters need to be estimated. Instead, CDO consistently provides enhanced predictive performances. We believe this may be due to the fact that, by imposing a threshold on the norm, A-optimality seeks only points that are far from the design’s center, without ensuring a distance between the data points that have already been collected. CDO, on the other hand, emphasizes points that correspond to a poor prediction, which is more likely associated with a design area that the learner has not thoroughly explored. As a result, we are less prone to acquire the labels of data points in locations where we have already collected a significant number of observations. It should be noted that in real-time applications the improvement offered by active learning is not as large as the one that can be obtained in offline scenarios, where we can deterministically maximize the desired optimality criterion over a closed set of observations. Moreover, by setting $\alpha=10\%$ we are not being too demanding in terms of selecting observations with large norms for the A-optimality or high prediction variances for CDO. In Figure \ref{fig:numsim2}, we try to widen the gap with the random strategy by lowering $\alpha$, in this case up to 0.01. By raising the threshold, we can be more demanding in terms of the desirability of the selected instances. The only drawback is that the algorithms will need to span more observations to achieve the desired size for the augmented design and meet the budget constraint. We believe this may not represent an issue since data is nowadays collected at very high sampling rates. However, in the final decision concerning the level of $\alpha$, practitioners will need to make a trade-off between the desired prediction improvement and the time required to select the new labeled examples.

Figure \ref{fig:numsim2} reports the learning curves obtained using a smaller $\alpha$. As expected, the enhancement obtained using the proposed strategy is increased with respect to the passive random sampling. However, it is worth noting that the improvement is more evident when the number of parameters is smaller, as the gain obtained in the high-dimensional cases was already significant with $\alpha=10\%$.

Finally, we analyze the computational time required by the two active learning strategies. To this extent, we introduce a measure called average decision time, which quantifies the time required to decide whether to query the label of an unlabeled observation or discard it. The results obtained on the numerical simulations, for different number of process variables, are reported in Table \ref{tab:comp}. Both active learning strategies are highly efficient and do not require a high computational time. According to the CDO strategy, at each iteration we are simply computing the UPV for the new data point, which requires less time than computing the norm of the new observation. It should be noted that the average decision time is lower because the inverse of the whitened moment matrix, $\left(\mathbf{Z}^\top \mathbf{Z}\right)^{-1}$, does not need to be computed at each iteration. However, it must be updated when the design is augmented by including an additional labeled observation. Updating and inverting the whitened moment matrix takes, on average, 0.31375 milliseconds (ms). 


\begin{table}
\caption{Average decision times (ms) for the two active learning methods (50 simulations).}
\label{tab:comp}
\centering
\begin{tabular}{ccccc}
\textbf{Strategy}    & \textbf{10 variables} & \textbf{20 variables} & \textbf{50 variables} & \textbf{100 variables} \\
\hline
CDO                  & 0.00494               & 0.00527               & 0.00568               & 0.00690                \\
Norm-Thresholding    & 0.00635               & 0.00642               & 0.00673               & 0.00716                \\
\hline
\end{tabular}
\end{table}

From an operational point of view, the average decision time is a highly relevant metric and it is closely related to the specific sampling frequency of the process. Indeed, to allow for a timely instance selection, the decision time should be strictly lower than the expiry date of the unlabeled data point, which is given by the time window where it is possible to query its label.

\subsection{Tennessee Eastman Process}

\begin{figure}[h]
  \centering
  \includegraphics[width=0.8\linewidth]{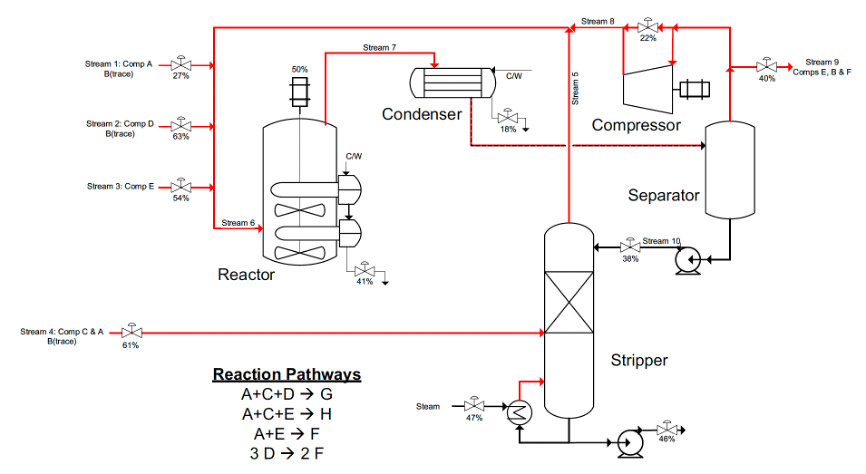}
  \caption{The TEP piping and instrumentation diagram \cite{Andersen2022}.}
  \label{fig:tep}
\end{figure}

The Tennessee Eastman Process (TEP) is a commonly used benchmark in industrial and chemical engineering research and it has been thoroughly investigated in terms of process dynamics and control \cite{Ricker1995,LawrenceRicker,McAvoy1994,Capaci2019,Lyman1995}. Recently, it has been also used to validate active learning or soft sensor modeling approaches \cite{Bao2015,Jia2020,Zhu2015,grbic,Yin2018}. It was initially published in 1993 \cite{Downs1993} but since then it has been further developed and improved. For this study, we used a recently released MATLAB simulator to generate the data \cite{Andersen2022,Reinartz2021}. We generated 50 datasets with the process running in normal operating conditions, using a sampling rate of approximately 1 minute. Figure \ref{fig:tep} depicts the TEP flowchart, which shows how the process is primarily composed of a reactor, a product condenser and separator, a stripper, and a compressor.

The TEP, like many other industrial processes, includes some easy-to-measure process variables whose real value can easily be monitored online, and some hard-to-measure variables, which are difficult to track during routine operations. Data-driven soft sensors are often developed to predict the latter in real-time. However, training regression models frequently necessitates a large number of labeled examples, and conducting quality inspections on chemical products may be costly and time-consuming. For this reason, optimizing the sampling strategy using active learning is highly desirable. The 16 process variables shown in Table \ref{tab:tep} are often used as predictors for the hard-to-measure process variables when testing active learning or soft sensor modeling approaches on the TEP. In most cases, the response variable is one of the composition measurements, such as the purge or product streams \cite{Bao2015,Zhu2015,grbic}. In this work, we selected two purge streams (Stream 9A and Stream 9E) and two product streams (Stream 11D and Stream 11E) as the response to be predicted using the easy-to-measure variables.


\begin{table}
\caption{Variables of the TEP used as predictors in the regression models.}
\label{tab:tep}
\centering
\begin{tabular}{ccc}
\textbf{Number} & \textbf{Process Variable}                  & \textbf{Code} \\
\hline
1               & A feed                                     & XMEAS1        \\
2               & D feed                                     & XMEAS2        \\
3               & E feed                                     & XMEAS3        \\
4               & A and C feed                               & XMEAS4        \\
5               & Recycle flow                               & XMEAS5        \\
6               & Reactor feed rate                          & XMEAS6        \\
7               & Reactor temperature                        & XMEAS9        \\
8               & Purge rate                                 & XMEAS10       \\
9               & Product separator temperature              & XMEAS11       \\
10              & Product separator pressure                 & XMEAS13       \\
11              & Product separator underflow                & XMEAS14       \\
12              & Stripper pressure                          & XMEAS16       \\
13              & Stripper temperature                       & XMEAS18       \\
14              & Separator steam flow                       & XMEAS19       \\
15              & Reactor cooling water outlet temperature   & XMEAS21       \\
16              & Separator cooling water outlet temperature & XMEAS22   \\
\hline
\end{tabular}
\end{table}

\begin{figure}[h]
  \centering
  \includegraphics[width=\linewidth]{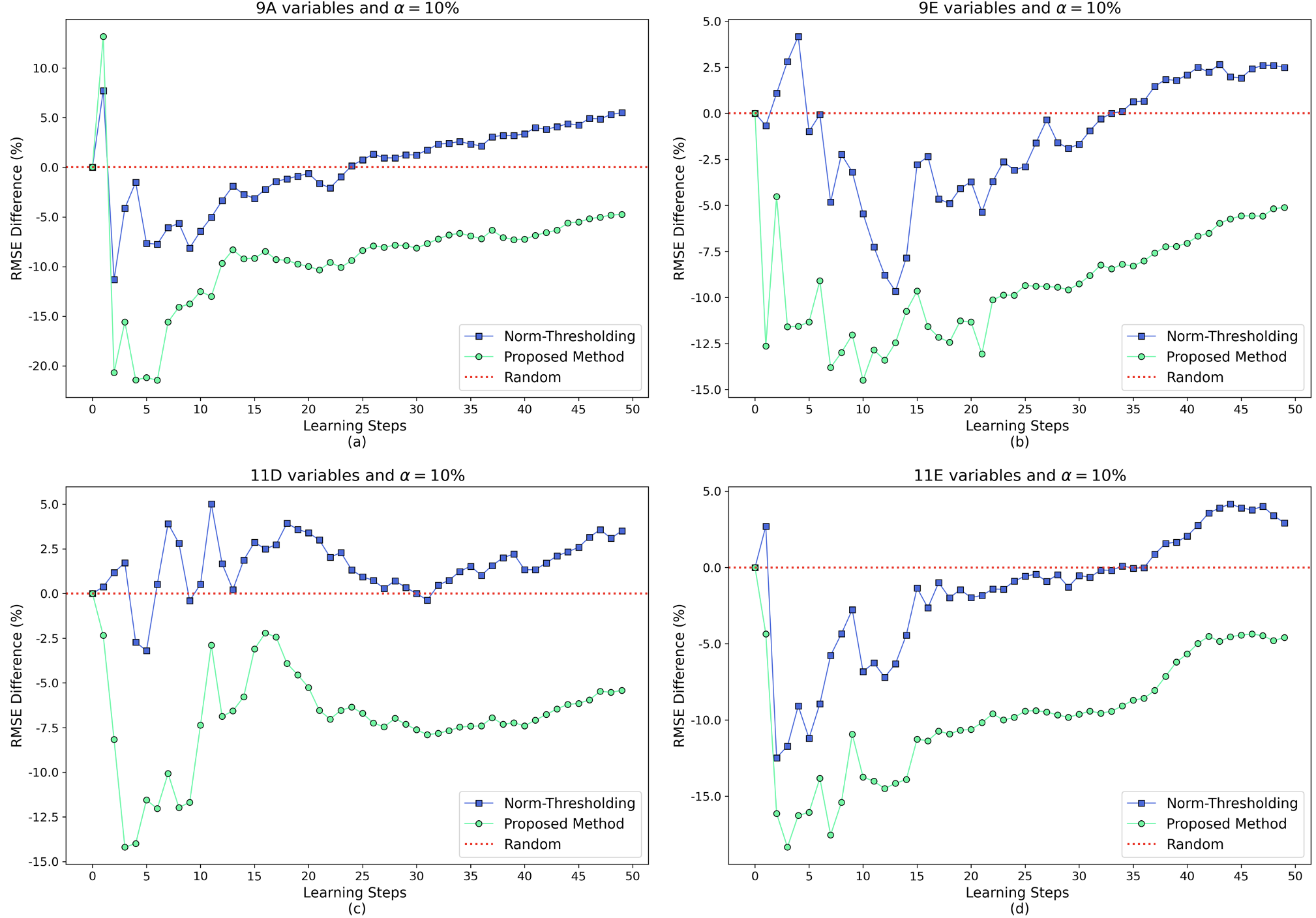}
  \caption{Percentage difference in RMSE between random sampling and the active learning methods, using $\alpha$=10\% (50 simulations).}
  \label{fig:tep1}
\end{figure}

\begin{figure}[H]
  \centering
  \includegraphics[width=\linewidth]{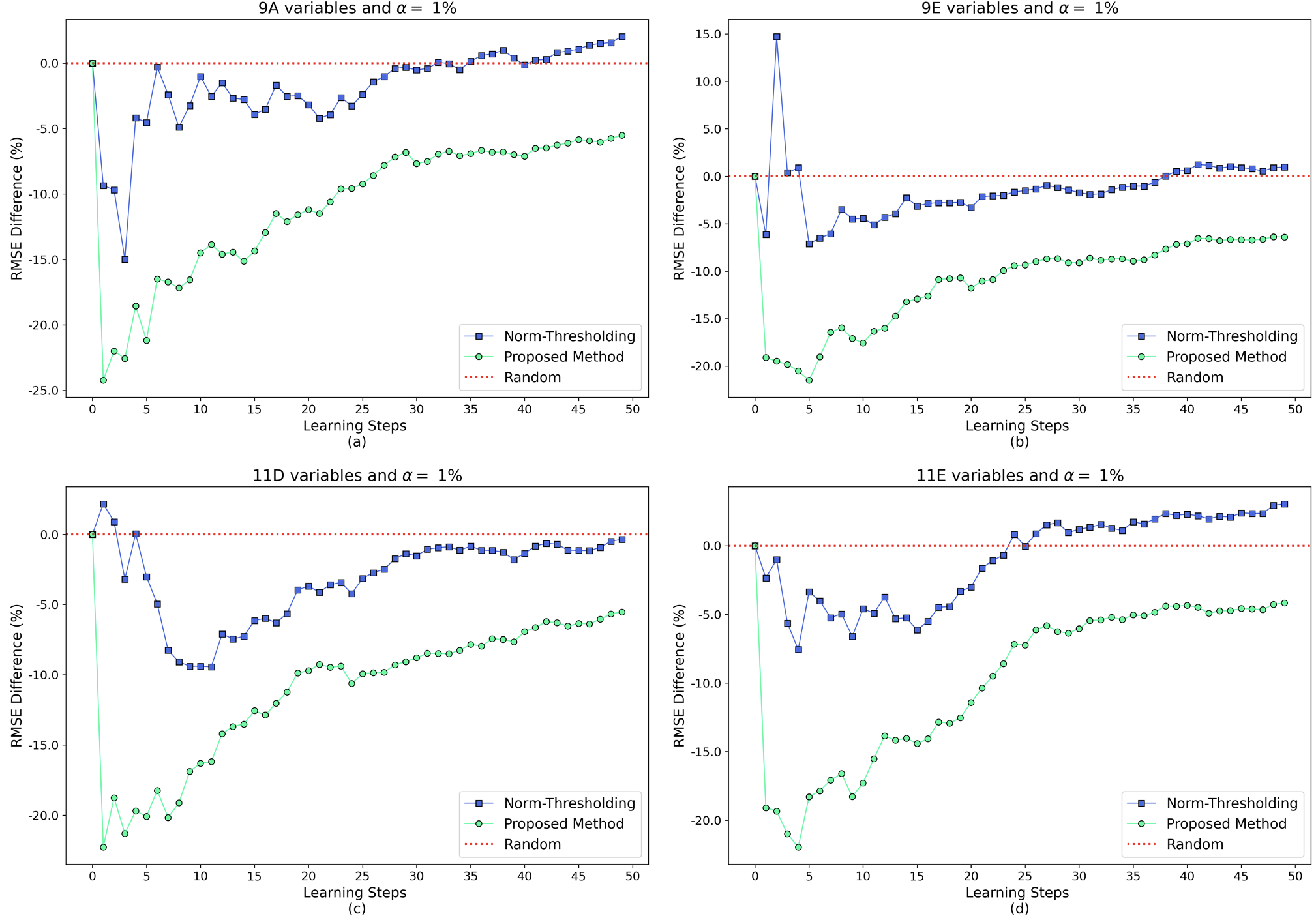}
  \caption{Percentage difference in RMSE between random sampling and the active learning methods, using $\alpha$=1\% (50 simulations).}
  \label{fig:tep2}
\end{figure}

As in the case of the numerical simulations, 50 datasets have been generated, and the average RMSE results are presented in the learning curve plots in Figures \ref{fig:tep1} and \ref{fig:tep2}. Most of the experimental parameters correspond to the ones used in the numerical study. The number of observations allocated to the first training set is equal to $p+2$, which in this case corresponds to 18. The warm-up length $w$ is equal to 500 and the budget $b$ is set to 50. The main difference from the models used in Section \ref{subsec:simulations} is that, in this case, all the models include the intercept term. We can see in Figure \ref{fig:tep1} how the results obtained in Section \ref{subsec:simulations} are still valid with data coming from a realistic industrial process simulator. Indeed, both the random and norm-thresholding approaches are outperformed by the proposed strategy. With regards to the level of $\alpha$, the behavior observed in the numerical study does not seem to be altered and, as the threshold is raised, the performance gap between random sampling and active learning strategies widens. 

\begin{figure}[H]
  \centering
  \includegraphics[width=\linewidth]{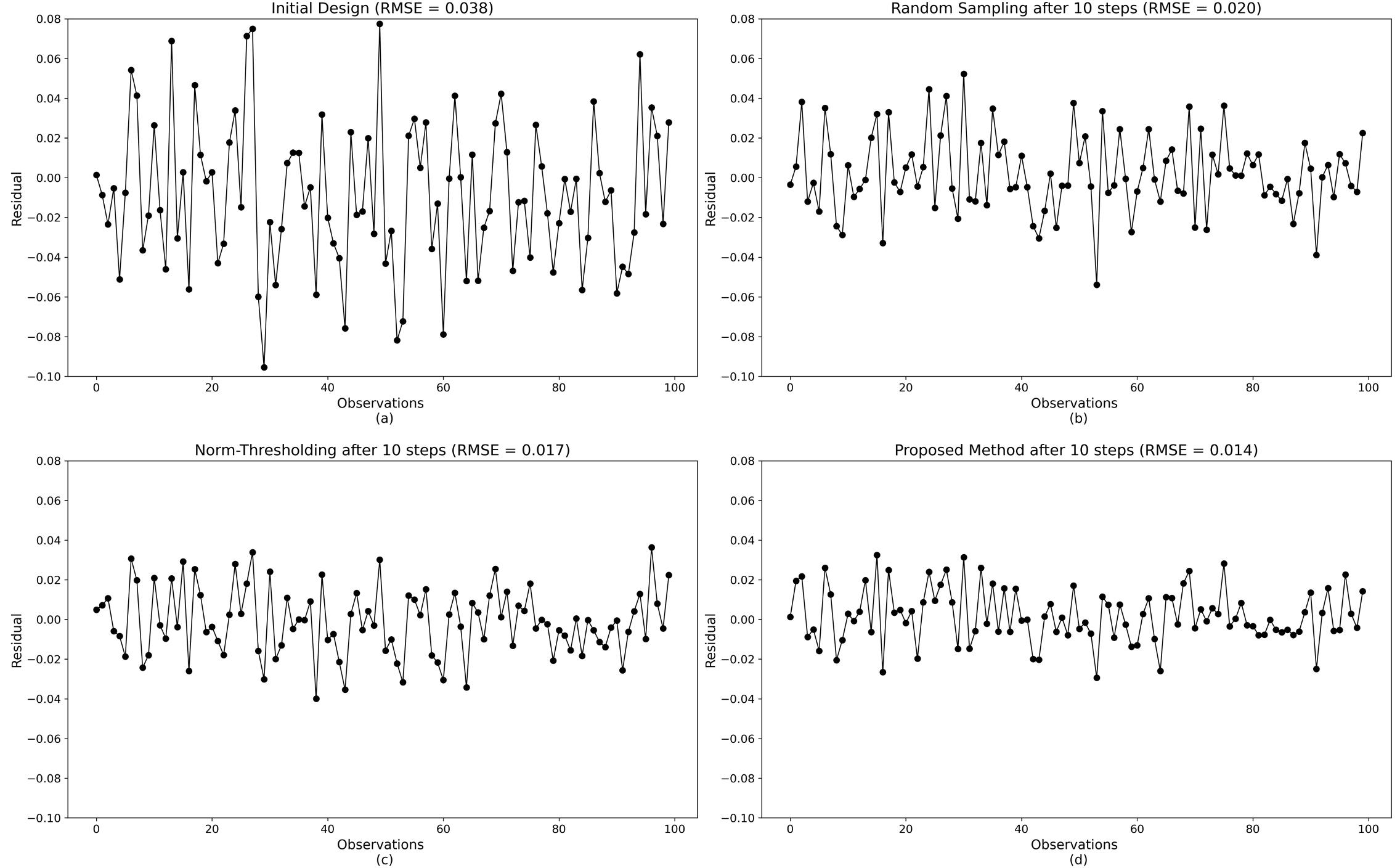}
  \caption{Residuals of the Stream 9A predictions: with the initial training set (a) and after augmenting the design with 5 additional labeled examples with the different methods (b-d) (one simulation with $\alpha=1\%$).}
  \label{fig:res1}
\end{figure}

The plots in Figure \ref{fig:res1} show the residuals related to the first composition measurements analyzed, stream A of the purge. For illustrative purposes, the residuals refer to a smaller test set, composed of 100 observations. The first plot (a) shows the residuals obtained with the first random design, which is common to all the compared approaches. The remaining plots (b-c) illustrate the residuals obtained after five learning steps with each strategy. In general, we can see how the predictive performance improves when more observations are included in the design. However, the predictions obtained with the proposed strategy are significantly better than the ones obtained with random sampling and norm-thresholding. Indeed, it should be noted how the RMSE obtained with the fifth model using CDO is 55 percent lower than the RMSE obtained with random sampling, and 23 percent lower than the RMSE obtained with the alternative active learning scheme. Finally, the improvement of CDO from the initial RMSE is higher than 65 percent. It should be noted how a simple linear regression model fitted on a small training set can achieve compelling prediction results when the labeled examples are appropriately selected. This is true even when testing our approach on data from the TEP, which is characterized by highly nonlinear relationships.

\begin{figure}[H]
  \centering
  \includegraphics[width=\linewidth]{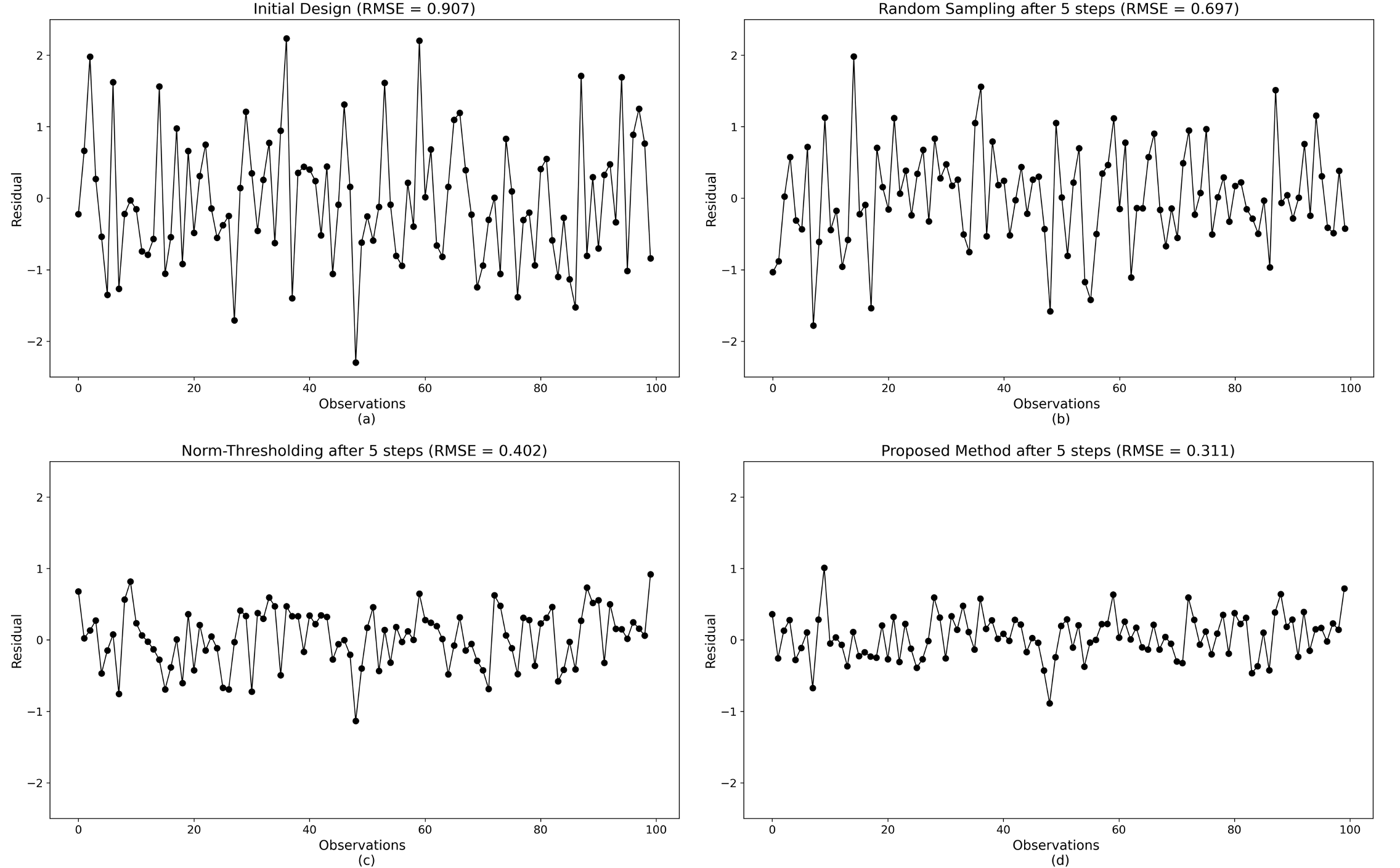}
  \caption{Residuals of the Stream 11D predictions: with the initial training set (a) and after augmenting the design with 10 additional labeled examples with the different methods (b-d) (one simulation with $\alpha=1\%$).}
  \label{fig:res2}
\end{figure}

Figure \ref{fig:res2} shows the predictions obtained for stream D of the product. In this case, to offer an additional view, we compared the models obtained after 10 learning steps. It can be seen how the behavior of the different schemes follows the same trend observed in Figure \ref{fig:res1}. Indeed, after 10 iterations, the RMSE obtained with CDO is 18 percent lower than the one obtained by norm-thresholding and 30 percent lower than the one obtained with random sampling. From the initial design, the RMSE is reduced by more than 60 percent with CDO.

\section{Conclusion} \label{sec:end}
In many industrial processes and real-life applications, data is often abundant only in an unlabeled form. Moreover, the prohibitive cost required by quality inspections and the time required by manual annotation makes it unfeasible to label each data point with its quality characteristic. In these cases, active learning can significantly improve the predictive performance of regression models by smartly selecting the instances to include in the training set. In situations where many observations are sequentially processed, it is necessary to provide a real-time sampling strategy for selecting the most informative instances. In this paper, we propose an optimal strategy for performing stream-based active learning with linear regression models. Two case studies, one using numerical simulations and the other one using the TEP, show that the proposed approach offers improved predictive performance and reduces the prediction error faster.




\newpage

\typeout{}
\bibliography{sn-bibliography}

\end{document}